\documentclass[letterpaper]{article} 
\usepackage{aaai23}  
\usepackage{times}  
\usepackage{helvet}  
\usepackage{courier}  
\usepackage[hyphens]{url}  
\usepackage{graphicx} 
\usepackage{natbib}  

\usepackage{caption} 
\usepackage[table,xcdraw]{xcolor}
\usepackage{algorithm}
\usepackage{algorithmic}
\usepackage{booktabs}
\usepackage{tabularx}

\usepackage{tikz}
\usepackage{xspace}
\RequirePackage{fontawesome}
\usepackage{color,soul} 
\usepackage{url}

\usepackage{subcaption}
\usepackage{graphicx}
\usepackage{multirow}
\usepackage[most]{tcolorbox}

\usepackage{paralist}
\usepackage[hidelinks]{hyperref}

\usepackage{enumitem}

\newcommand{\hilightw}[1]{\setlength{\fboxsep}{1pt}\colorbox{red}{#1}}
\newcommand{\hilightc}[1]{\setlength{\fboxsep}{1pt}\colorbox{green}{#1}}

\newcommand{\hlitemw}{\stepcounter{enumi}\item[\hilightw{\theenumi}]}
\newcommand{\hlitemc}{\stepcounter{enumi}\item[\hilightc{\theenumi}]}

\newcommand{\datasetname}{\texttt{GameBugDescriptions}\xspace}

\usepackage[framemethod=TikZ]{mdframed}

\definecolor{sampleblue}{rgb}{0.62, 0.77, 0.9}
\definecolor{llmocolor}{rgb}{0.62, 0.77, 0.9}

\newcommand{\llmoutput}[1]{{\sethlcolor{llmocolor}\hl{#1}}}

\newcounter{davinci}[section]

\newenvironment{davinci}[2][]{%
\refstepcounter{davinci}%
\ifstrempty{#1}%
{\mdfsetup{%
frametitle={%
\tikz[baseline=(current bounding box.east),outer sep=0pt]
\node[anchor=east,rectangle,fill=sampleblue]
{\strut Sample~\theoptxlarge~  -- ~\texttt{text-davinci-002}  } ;}}
}%
{\mdfsetup{%
frametitle={%
\tikz[baseline=(current bounding box.east),outer sep=0pt]
\node[anchor=east,rectangle,fill=sampleblue]
{\strut \texttt{text-davinci-001}- Sample ~\thetheo:~#1};}}%
}%
\mdfsetup{innertopmargin=0pt,linecolor=sampleblue,%
linewidth=2pt,topline=true,%
frametitleaboveskip=\dimexpr-\ht\strutbox\relax
}
\begin{mdframed}[]\relax%
\label{#2}}{\end{mdframed}}

\newcounter{optxlarge}[section]

\renewcommand{\theoptxlarge}{\arabic{davinci}}
\newenvironment{optxlarge}[2][]{%
\refstepcounter{davinci}%
\ifstrempty{#1}%
{\mdfsetup{%
frametitle={%
\tikz[baseline=(current bounding box.east),outer sep=0pt]
\node[anchor=east,rectangle,fill=sampleblue]
{\strut Sample~\theoptxlarge~  -- ~\texttt{OPT-175B} };}}
}%
{\mdfsetup{%
frametitle={%
\tikz[baseline=(current bounding box.east),outer sep=0pt]
\node[anchor=east,rectangle,fill=sampleblue]
{\strut text-optxlarge-001~\thetheo:~#1};}}%
}%
\mdfsetup{innertopmargin=0pt,linecolor=sampleblue,%
linewidth=2pt,topline=true,%
frametitleaboveskip=\dimexpr-\ht\strutbox\relax
}
\begin{mdframed}[]\relax%
\label{#2}}{\end{mdframed}}

\definecolor{ModelNameColor}{rgb}{0.94, 0.87, 0.68}

\usepackage{newfloat}
\usepackage{listings}
\DeclareCaptionStyle{ruled}{labelfont=normalfont,labelsep=colon,strut=off} 
\floatstyle{ruled}
\newfloat{listing}{tb}{lst}{}
\floatname{listing}{Listing}
\usepackage[T1]{fontenc}

\newcommand{\papertitle}{Large Language Models are Pretty Good Zero-Shot Video Game Bug Detectors}
\title{\papertitle}
\author{
}

\author{%
Mohammad Reza Taesiri \quad  Finlay Macklon \quad Yihe Wang \quad Hengshuo Shen \\ Cor-Paul Bezemer \\
University of Alberta \\
\texttt{\{taesiri,macklon,yihe2,hengshuo,bezemer\}@ualberta.ca}\\
}

\usepackage{bibentry}

\begin{document}

\maketitle

\begin{abstract}
Video game testing requires game-specific knowledge as well as common sense reasoning about the events in the game. While AI-driven agents can satisfy the first requirement, it is not yet possible to meet the second requirement automatically.
Therefore, video game testing often still relies on manual testing, and human testers are  required to play the game thoroughly to detect bugs. As a result, it is challenging to fully automate game testing.
In this study, we explore the possibility of leveraging the zero-shot capabilities of large language models for video game bug detection.
By formulating the bug detection problem as a question-answering task, we show that large language models can identify which event is buggy in a sequence of textual descriptions of events from a game. To this end, we introduce the \datasetname benchmark dataset, which consists of 167 buggy gameplay videos and a total of 334 question-answer pairs across 8 games.
We extensively evaluate the performance of six models across the OPT and InstructGPT large language model families on our benchmark dataset.
Our results show promising results for employing language models to detect video game bugs. With the proper prompting technique, we could achieve an accuracy of 70.66\%, and on some video games, up to 78.94\%. Our code, evaluation data and the benchmark can be found on \url{https://asgaardlab.github.io/LLMxBugs}
\end{abstract}


\section{Introduction}



Similar to other software products, a video game must be thoroughly tested to assure its quality. Game testing is an umbrella term for many types of tests that cover different aspects of the game. For example, a rendering test aims to verify the visual quality of the output, whereas a gameplay test assesses whether the game is engaging enough. While it is possible to automate some game testing elements, e.g., by using advanced vision models to detect graphical issues automatically~\cite{taesiri2022clip}, most game testing aspects still require a human tester~\cite{pascarella2018video}. Two of the main challenges that prevent the automation of game testing are the difficulty to automate (1)~knowledge about the game context and (2)~common sense reasoning~\cite{politowski2021survey}. 

\begin{figure}[!t]
    \centering
    \includegraphics[width=1\columnwidth]{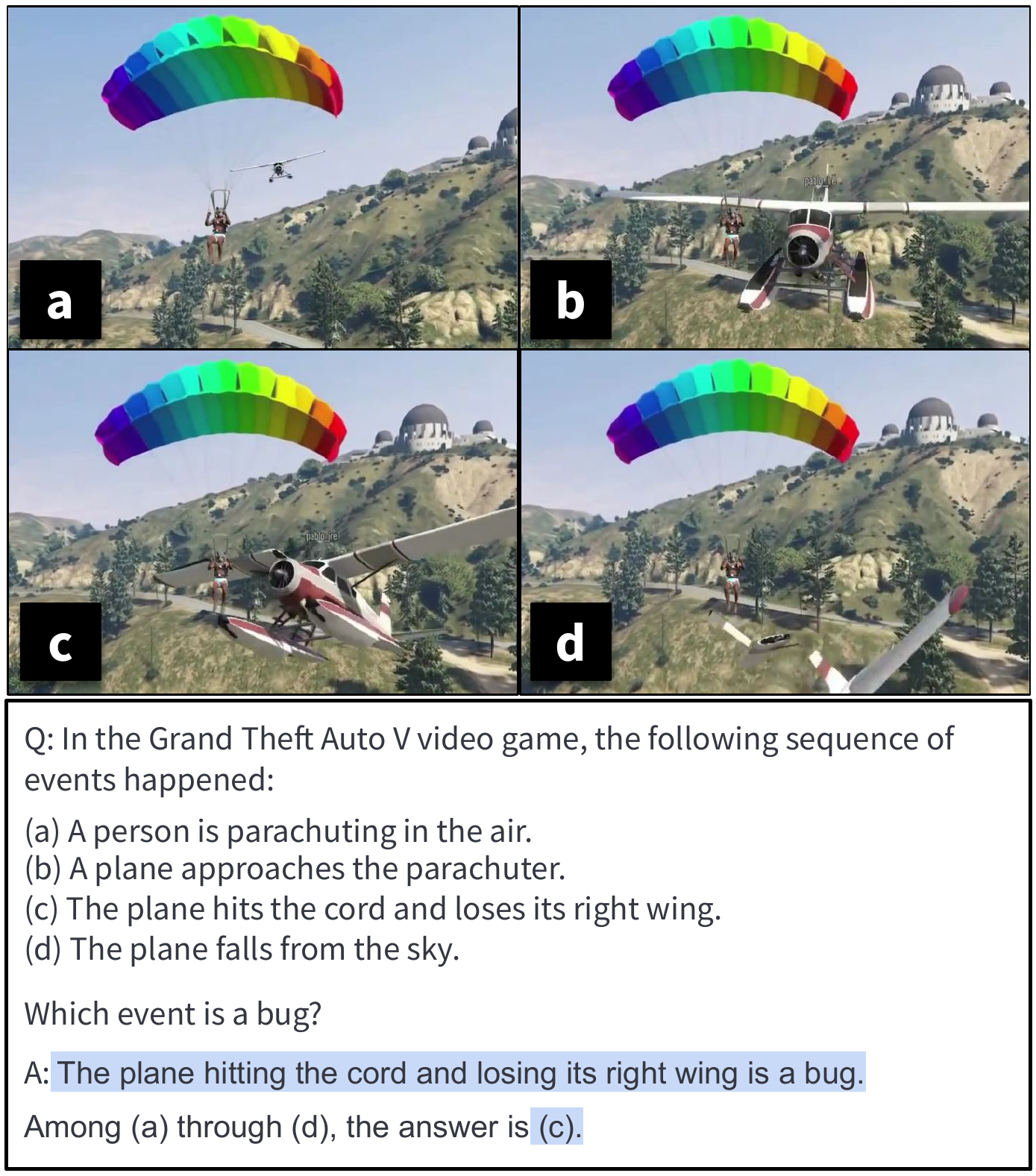}
    \caption{An example of using a large language model to detect a video game bug by classifying a sequence of events in the \textit{Grand Theft Auto V}\protect\footnotemark  \; video game in which a collision between a plane and parachute cords leads to the plane losing its right wing. The highlighted text shows the response of the davinci model from the InstructGPT family.}
    \label{fig:example_detector}
\end{figure}

\footnotetext{Source: \url{https://redd.it/2s5xon}}

Many video games rely on a physics engine that defines the rules of the world in which the game is situated~\cite{millington2007game}. For some games, there are sharp contrasts between the game world and the natural laws of the real world. These differences make it hard to reason about events in video games without knowing the game context. For example, is it a bug that the player survives after falling from a very high height? Answering such a question is impossible without having knowledge about the target video game. 

In this study, we propose using the game context knowledge and common sense reasoning capabilities of large language models (LLMs) to identify buggy events in video games and classify their bug type.
Recent revolutions in natural language processing (NLP) show that scaling up language models is beneficial in many tasks \citep{vaswani2017attention, devlin2018bert, rae2021scaling,  chowdhery2022palm, thoppilan2022lamda}, such as few-shot and zero-shot learning \citep{brown2020language, kojima2022large}.
Having been trained on very large datasets, LLMs have the potential to capture many details about topics in their training set, including video games.
Figure~\ref{fig:example_detector} shows an example of successful bug detection by a language model. 

We are the first to empirically evaluate the capability of LLMs as zero-shot video game bug detectors. Our main contributions are as follows:

\begin{enumerate}
    \item We present the \datasetname  dataset, the first dataset of videos of game bugs with a step-by-step textual description for bug detection purposes. This dataset can serve as an out-of-distribution (OOD) challenge for LLMs.
    \item We are the first to show that large language models have promising capabilities to detect video game bugs.
    \item We extensively evaluate the performance of two families of large language models on the bug detection and bug type classification tasks: InstructGPT \cite{ouyang2022training} and OPT  \cite{zhang2022opt}.
    \item We analyze the robustness of language models to different descriptions of the same event for these tasks.
\end{enumerate}

Our study demonstrates the promising capabilities of LLMs to play an important role in the automation of the game testing process.





\section{Background and Related Work}
Our work bridges the language modeling, video game, and software engineering research communities. In this section, we provide a brief overview of the relevant literature across these disciplines, in particular, on large language models and prompt engineering, and automated game testing.

\subsection{Large Language Models and Prompt Engineering}
The training objective in a language model is to learn a probability distribution over some text corpus. Such a simple training objective combined with sufficient model scaling can yield large language models that are successful even for tasks for which the model was not explicitly trained ~\citep{kaplan2020scaling, brown2020language, chowdhery2022palm, thoppilan2022lamda}.


Prompting or prompt engineering \citep{liu2021pre} is an effective technique wherein we condition a language model on a set of manually handcrafted  \citep{schick2020s, kojima2022large} or automated \citep{gao2020making} templates to solve new tasks. 
That is, new tasks can be solved by giving natural language instructions to a pretrained model without any further training, e.g., by providing sample reasoning steps to the model \cite{wei2022chain} in a few-shot setting.
Moreover, \citet{kojima2022large} showed that even with a prompting technique as simple as adding \textit{``Let's think step by step''} to the beginning of the answer, it is possible to trigger the reasoning in language models, which leads to higher accuracy improvement on multiple benchmarks in zero-shot setting.
Using graphical models \citet{dohan2022language} introduced a general formulation for prompted models, enabling  probabilistic programming with LLMs.

Several successful applications of LLMs include program synthesis~\cite{jain2022jigsaw}, code generation~\cite{chen2021evaluating} or chatbots~\cite{thoppilan2022lamda}. However, we are the first to apply LLMs to detect bugs in video games.


\subsection{Automated Game Testing}
As shown by prior work, automated game testing is challenging because game-specific knowledge and common sense reasoning are required to detect and report bugs~\cite{pascarella2018video, politowski2021survey}.
The majority of prior work on automated game testing focuses on methods to automatically play games, such as heuristic search strategies~\cite{keehl2019monster}.
Automated play techniques using reinforcement learning or evolutionary strategies ~\cite{zheng2019wuji, vinyals2019grandmaster, berner2019dota, justesen2019deep} allow the testing of video games from different perspectives, such as playability, game balance, and even predicting user churn rate \cite{roohi2020predicting, roohi2021predicting}.
However, these methods are often designed to maximize a certain reward function, which might lead to progress in the game in an unintended manner and even break the game's rules or physics engine \cite{baker2020emergent, clark_amodei_2019}.
More importantly, these methods do not have common sense reasoning.

Other prior work has leveraged computer vision and NLP techniques for automated video game testing. Several studies have proposed approaches for graphical bug detection using deep learning~\cite{ling2020using,taesiri2020video} or digital image processing~\cite{macklon2022automatically}. However, these approaches do not require common sense reasoning. For example, \citet{macklon2022automatically} rely on graphical assets of the game as a test oracle.

Several other studies have proposed approaches to retrieve moments from gameplay videos based on text queries~\cite{zhang2019retrieving,taesiri2022clip}. However, to detect bugs with these approaches the bug instance must be known in advance, and therefore these gameplay event retrieval approaches do not allow for automated detection of (previously undiscovered) bugs. Our approach does not have this requirement and can therefore be used to identify previously undiscovered bugs.

Finally, prior work has proposed NLP-based approaches to automatically improve test case descriptions for manual playtesting of games~\cite{viggiato2022identifying, viggiato2022using}, but we are the first to leverage LLMs for bug detection in video games.


\section{Bug Detection with Large Language Models} \label{sec:bugdetection}

To automatically identify buggy events in a video game, we propose using LLMs to reason about sequences of textual descriptions of game events. We formulate the problem as a question-answering (Q\&A) task \cite{srivastava2022beyond} for LLMs.
Here, we explain how we convert textual descriptions of a sequence of game events into a multiple-choice question and use a language model to identify the buggy event. In addition, we discuss how LLMs can assist us to classify the bug type. First, we motivate our methodology by demonstrating the game knowledge of LLMs.

\begin{figure}[!t]
    \centering
    \includegraphics[width=\linewidth]{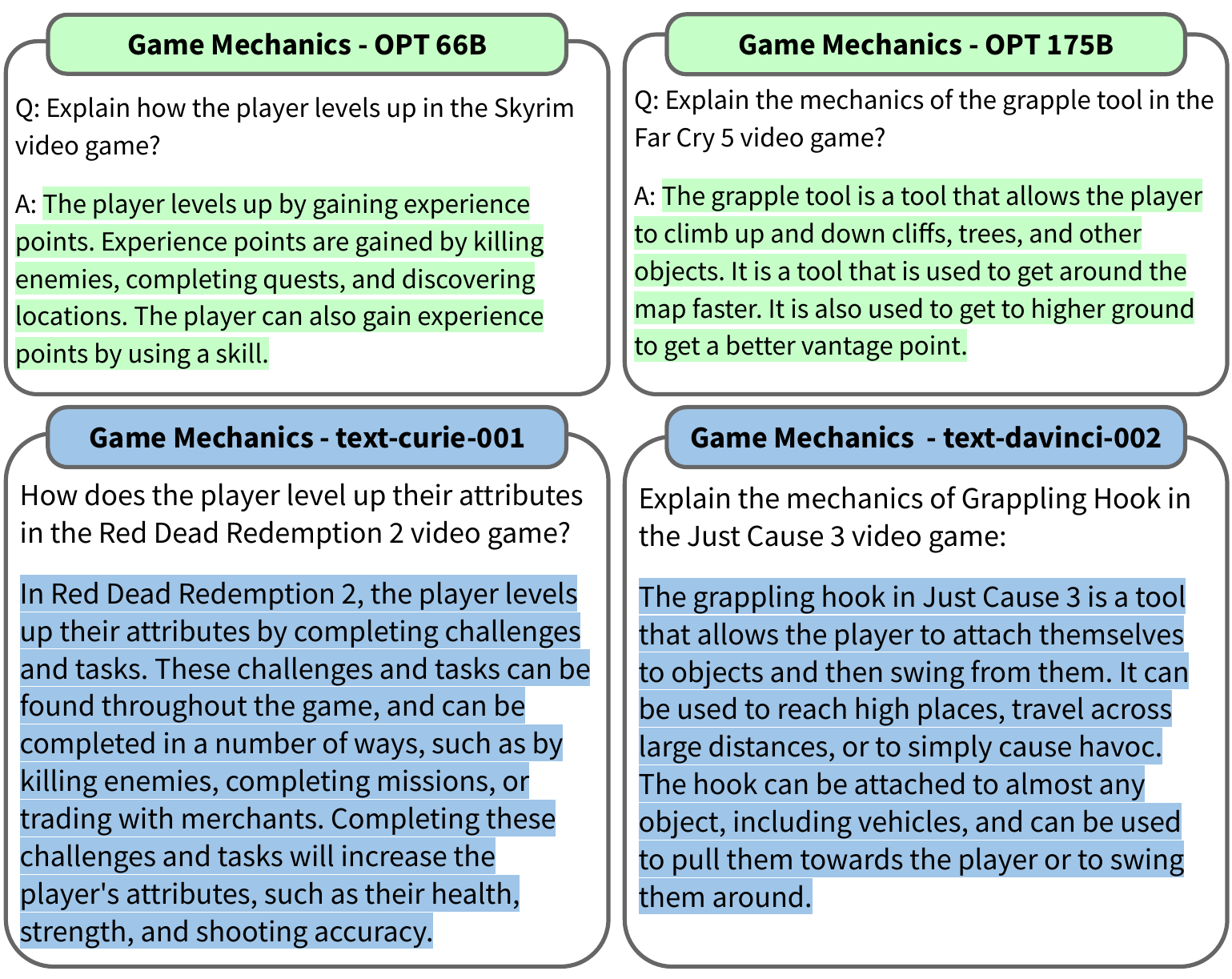}
    \caption{Examples of the game knowledge of LLMs.}
    \label{fig:kb}
\end{figure}

\paragraph{Large Language Models and Game Knowledge:}
To verify whether LLMs possess game knowledge, we conducted an informal analysis on a set of 34 multiple-choice and 10 free format questions that was designed by us. The davinci model can answer 91\% of the questions correctly.
Figure \ref{fig:kb} depicts a few cases showing language models of different sizes exhibiting game-specific knowledge in a zero-shot setting. In the remainder of this section, we explain how we leverage the game knowledge of LLMs for the more challenging task of bug detection for video games.

\paragraph{Problem Formulation:}  For a sequence $S$ of textual descriptions of events $e_i$ in a video game $G$ ($S=\{e_1, e_2, \cdots e_n\}$) we design the triplet $\langle Q, S, A \rangle$, in which $Q$ is a question template  and $A$ is the correct answer (i.e., the buggy event). By presenting $S$ as multiple choice options (e.g.,  \textbf{(a)} through \textbf{(d)} in Figure~\ref{fig:example_detector}), this triplet allows us to use a language model to detect bugs in a game event sequence. 

We leverage $Q$ to provide context information to the language model. Identifying buggy events in a video game requires context, because events in a video game can be unrealistic yet valid for that game.
For example, in the \textit{Grand Theft Auto V} video game, a player would die when they fall from a height, while in the \textit{Marvel's Spider-Man} game,  the player would not take any damage.
Therefore, we modify $Q$ to include a string template $G$, which is a placeholder for the game's name, which is used by the language model as a reference point. 
We design each question as follows:

\vspace{0.1cm}







\begin{tcolorbox}[colback=white]
\fontsize{8pt}{8pt}\selectfont{ \sffamily{In the [$G$] video game, the following sequence of events happened:\par}}

\vspace{0.05cm}

\normalsize{[$S$]}

\vspace{0.15cm}

\fontsize{8pt}{8pt}\selectfont{ \sffamily{ Which event is a bug? }}
\end{tcolorbox}







\subsection{Multi-Stage Prompting}



Following \citet{kojima2022large}, we propose a two-stage prompting technique for eliciting reasoning in language models. This technique enables LLMs to provide step-by-step reasoning for their answer, leading to higher accuracy and better interpretability, which allows us to debug the final answer and understand in which step the model goes wrong.

\paragraph{Stage 1 - Elicit Reasoning:}
Similar to prior work \citep{mccann2018natural,radford2019language,schick2021exploiting} for each triplet $\langle Q, S, A \rangle$, we use a string template ``\texttt{Q: [X]. [Z]}"  in which \texttt{[X]} is the input slot for concatenation of a question template $Q$ containing the game name, and the sequence of events $S$. The ``\texttt{Q:}" text in the template is a fixed string that implies we are asking a question. The \texttt{[Z]} slot is for the trigger sentence added to the beginning of the answer. The purpose of the trigger sentence is two-fold: (1)~to assist the language model in reaching the answer progressively, and (2)~to inject different perspectives in the classification, e.g., to see events as a game designer, player, or compared with the real world.

We feed the question and trigger sentence to a language model to provoke step-by-step reasoning and use the generated text in Stage 2.


\paragraph{Stage 2 - Answer Extraction:}
We rely on the language model to extract the final answer from the generated text, given the intermediate results, instead of performing text processing.
To this end, we concatenate the inputs and results from the previous stage and append an `answer extraction' prompt to the end. We can use different answer extraction prompts depending on the format of the final answer. Since we have a multiple choice format, we use the \textit{``Among (a) through (d), the answer is''} prompt.




\subsection{Bug Type Classification}



In addition to identifying the buggy event, classifying the type of bug is helpful for bug prioritization, as   described by \citet{truelove2021we}.
%
Therefore, as a secondary objective, we are interested in determining if language models can correctly classify video game bug types. To this end, we design another question template in which we provide the event description and the ground truth and ask the model to classify the type of the video game bug.

\vspace{0.1cm}






\begin{tcolorbox}[colback=white]
\fontsize{8pt}{8pt}\selectfont{ \sffamily{In the [$G$] video game, the following sequence of events happened:}}

\vspace{0.05cm}

\normalsize{[$e_{b}$]}

\vspace{0.1cm}

\fontsize{8pt}{8pt}\selectfont{ \sffamily{ What is the type of bug?}}

\vspace{0.05cm}

\normalsize{[$O$]}
\end{tcolorbox}

In which $O$ contains the bug types as multiple choice options, and $e_b$ is the buggy event in the original sequence.


\section{Dataset}

To evaluate the capabilities of LLM for identifying game bugs, we created the \datasetname dataset, a collection of videos of real game bugs. Our dataset consists of a collection of 167 buggy gameplay videos, each with 2 textual descriptions of the events in the videos, and a bug-type label per video. Each description includes up to four sentences describing the events in the video without reasoning about their bugginess. To assess the robustness of language models to different descriptions, we provide two descriptions written by two authors for each video. Each description is converted into a question-answering format as explained above, resulting in 334 question-answer pairs. We also classify all 334 descriptions into 9 different bug types.


\paragraph{Data Collection:} 

We start with the \texttt{GamePhysics} dataset \cite{taesiri2022clip}, a collection of gameplay videos from different games. Following prior work \cite{taesiri2022clip}, we focus on eight popular video games (see Table~\ref{tab:results_davinci}) from this dataset and sample a total of 200 videos.
We define several exclusion criteria to manually filter out videos that are unsuitable for our study: (1)~The video does not showcase a bug (but instead, e.g., a funny moment or impressive playing skills), (2)~The video showcases a severe graphical bug (e.g., a glitch) and (3)~The video is related to a game modification (e.g., the contents or logic of the game were changed through manual modification of the game files).



\paragraph{Labeling:}

After filtering, our dataset contains 167 videos, with an average of 20 videos per game. Two of the authors labelled each video separately without exchanging information during the labelling.\footnote{Both authors are fluent in English.}
Each label contains step-by-step textual descriptions of events in the video without interpretation or reasoning about the events.\footnote{These authors had no access to any language model during the labelling process and did not modify their labels in any way.}
The resulting dataset contains a total of 334 descriptions with an average of 3.9 sentences per video. In the rest of the paper, we denote each set of the descriptions as \textit{Descr1} and \textit{Descr2}.

We also provide one bug type for each video in our dataset. 
One of the authors manually classified each video into a single bug type, and another author finalized the classification by combining similar bug types while confirming that each video was suitably classified.
During the classification process, nine different bug types\footnote{Note that several game bug taxonomies exist, however, they are not detailed enough to showcase the capabilities of LLM. The purpose of our paper is not to create a new bug taxonomy, but instead to demonstrate these capabilities.} were extracted: 


\begin{figure}[!t]
    \centering
    \includegraphics[width=\columnwidth]{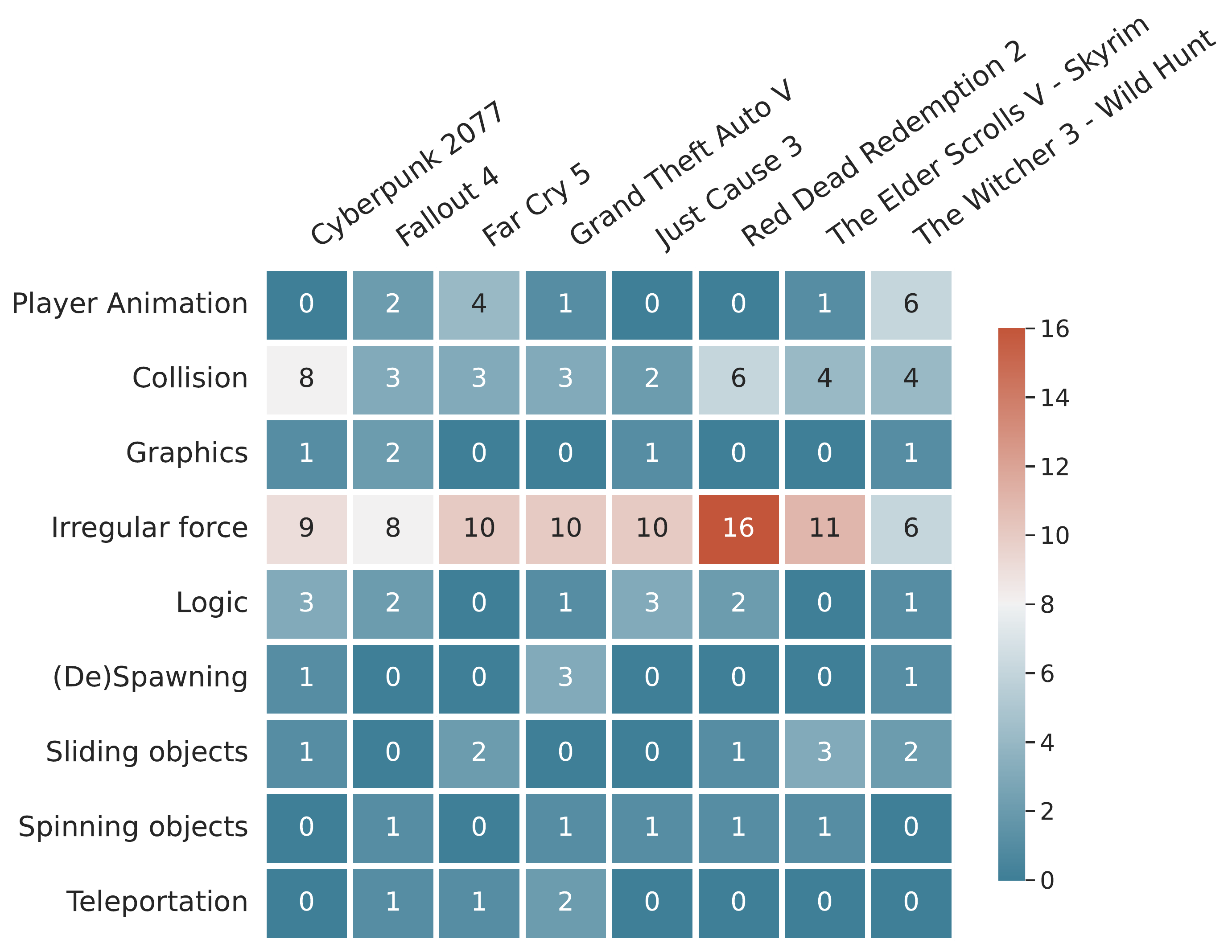}
    \caption{Distribution of bug types across games in the \datasetname dataset.}
    \label{fig:types}
\end{figure}


\begin{compactitem}
    \item \textbf{Player Animation:} When the player's body or limb is animated incorrectly, such as twisted to an abnormal angle or moving uncontrollably. 
    \item \textbf{Teleportation:} An instantaneous movement of an object or character from one point to another.
    \item \textbf{Graphics:} Objects or backgrounds are displayed incorrectly on the screen, e.g., showing the wrong color, texture, or with the screen flickering.
    \item \textbf{(De)Spawning:} An object suddenly appears in (or disappears out of) sight.
    \item \textbf{Collision:} Objects are clipping through each other, or objects have interactions they are not supposed to have, like hitting an invisible wall.
    \item \textbf{Spinning objects:} An object is rotating without force being exerted on it.
    \item \textbf{Sliding objects:} An object is sliding on the ground as if there is little or no friction.
    \item \textbf{Irregular force:} An object is acted upon by a force in a way that disobeys the game's physics, such as an object suddenly starting to float.
    \item \textbf{Logic:} Non-player characters (NPCs) do things that seem illogical, such as an ambulance driving on the sidewalk or running over pedestrians.
\end{compactitem}

The \textit{Irregular force} type is the most common bug type among all videos. Figure \ref{fig:types} shows the distribution of bug types in \datasetname dataset. 

\paragraph{Average CLIP Score:}  To estimate the similarity between the descriptions and videos (and hence the quality of the descriptions), we calculated the CLIP Score~\cite{radford2021learning} between each sentence in the event description and all frames in the video. We record the maximum similarity among frames for each sentence and report the average score for all sentences as a final score. The median CLIP Score for all 334 video descriptions in our dataset is 0.30, which has been previously used as a reasonable similarity threshold~\cite{schuhmann2021laion} based on human inspection.

\section{Experimental Setup}
\label{sec:exp_setup}
We executed two tasks in our experiments: (1)~buggy event identification and (2)~bug type classification. In this section, we discuss the experimental setup.

\paragraph{Models:}
We tested six models from the InstructGPT  \citep{ouyang2022training} and OPT \citep{zhang2022opt} families of models. We tested all four InstructGPT models (ada, babbage, curie and davinci), which contain 0.3B to 175B parameters\footnote{\url{https://blog.eleuther.ai/gpt3-model-sizes/}}. We also ran our experiments on OPT models with 66B and 175B parameters. In all experiments, we set the \texttt{temperature} parameter to 0, and the stopping sequence to ``Q:'' to prevent repetition. We set \texttt{max\_tokens} to 256 in the first stage and 32 in the answer extraction stage. For the InstructGPT models, we used the OpenAI API, and for the OPT models, we used the official implementation\footnote{\url{https://github.com/facebookresearch/metaseq/}}, which provides a similar API to what OpenAI offers. We hosted our OPT models on an NVIDIA DGX System with 8xA100 GPUs (80GB) and 2 TB of system memory.



\paragraph{Trigger Sentences:}

\begin{table}[!t]
\centering
\caption{The used trigger sentences.}
\label{tab:triggers}
\resizebox{\columnwidth}{!}{%
\begin{tabular}{@{}ll@{}}
\toprule
\textbf{\#} & \textbf{Trigger Sentence} \\ \midrule
1 & \textit{``Let's reason the events according to the reference game''} \\
2 & \textit{``Let’s think step by step.''} \\
3 & \textit{``According to the rules of the game''} \\
4 & \textit{``The reference game is''} \\
5 & \textit{``Let's think like a game tester''} \\
6 & \textit{``First,''} \\
7 & \textit{``Let's think like a game designer.''} \\ \bottomrule
\end{tabular}}
\end{table}

We tested a total of seven  trigger sentences, including two top-performing ones from prior work \cite{kojima2022large}. In particular, we included the trigger sentence \textit{``Let's think step by step''}, because it has been shown to boost the performance of the davinci variant of InstructGPT across many tasks~\cite{kojima2022large}. We also included \textit{``First,''} as a simple baseline~\cite{ahn2022can}. We added five new trigger sentences to dictate different viewpoints and enforce the game's rules to the language model. The complete list of trigger sentences can be seen in Table~\ref{tab:triggers}.

\paragraph{Bug Type Classification:}
In the bug type classification task, we only feed the question and expect the model to choose the correct bug types without the use of any triggering sentences.

\paragraph{Evaluating the Experiments:}
For the buggy event identification task, we calculated the accuracy for each set of descriptions separately, and report the average for all combinations of language models and trigger sentences. We also calculated the accuracy per game for the top performing model. To estimate the robustness of the buggy event identification under different descriptions of the same sequence of events, we used the Wilcoxon signed-rank test~\cite{woolson2007wilcoxon} to determine if there was a statistically significant difference between accuracies for each set of descriptions with the top performing model. For the bug type classification task, we determined accuracy both per bug type and on the entire dataset. We discuss the correctness of the reasoning of the models in the Discussion section.


\section{Results}
\begin{table}[!t]
\centering
\caption{Breakdown of bug type classification accuracy of the davinci model (in \%)}
\label{tab:bugtype_accuracy}
\begin{tabular*}{\columnwidth}{l@{\extracolsep{\fill}}r@{\extracolsep{\fill}}r}
\toprule
\multicolumn{1}{l}{\textbf{Bug Type}} & \multicolumn{1}{r}{\textbf{Count}} & \multicolumn{1}{r}{\textbf{Accuracy}} \\ \midrule
\textit{Player Animation} & 28 & 96.43 \\
\textit{Teleportation} & 8 & 75.00 \\
\textit{Graphics} & 10 & 70.00 \\
\textit{(De)Spawning} & 10 & 70.00 \\
\textit{Collision} & 66 & 69.70 \\
\textit{Spinning objects} & 10 & 50.00 \\
\textit{Sliding objects} & 18 & 33.33 \\
\textit{Irregular force} & 160 & 24.38 \\
\textit{Logic} & 24 & 16.67 \\ \midrule
\textit{\textbf{Average} (total)} & 334 & 44.01 \\ 
\bottomrule
\end{tabular*}
\end{table}

\subsection{Task 1: Buggy Event Identification}
LLMs show promising zero-shot performance for buggy event identification on our dataset. Although the accuracy varies depending on the model size and trigger sentence, our results suggest that LLMs can be utilized for game bug detection tasks. Table~\ref{tab:main_table} shows the accuracy of the models across the various trigger sentences and descriptions.

The davinci model delivers the best performance, and it can achieve up to 70.66\% accuracy using the first set of descriptions (\texttt{Descr1}) and the trigger sentence \textit{``Let's reason the events according to the reference game''}. The accuracy averages 65.27\% across the entire dataset.

We find there are no consistent trends among different models. For example, for the OPT-66 model, the trigger sentence \textit{``According to the rules of the game''} leads to the best average performance (40.12\%).

\begin{table*}[!t]
\centering
\caption{Breakdown of buggy event identification accuracy by descriptions, models and trigger sentences -- Bold numbers show the highest performing model and trigger sentence combination (in \%). }
\label{tab:main_table}
\resizebox{\textwidth}{!}{%
\begin{tabular}{@{}c
>{\columncolor[HTML]{D9D9D9}}r 
>{\columncolor[HTML]{D9D9D9}}r 
>{\columncolor[HTML]{D9D9D9}}r rrr
>{\columncolor[HTML]{D9D9D9}}r 
>{\columncolor[HTML]{D9D9D9}}r 
>{\columncolor[HTML]{D9D9D9}}r rrr
>{\columncolor[HTML]{D9D9D9}}r 
>{\columncolor[HTML]{D9D9D9}}r 
>{\columncolor[HTML]{D9D9D9}}r rrr@{}}
\toprule
 & \multicolumn{3}{c}{\cellcolor[HTML]{D9D9D9}\textbf{OPT-66B}} & \multicolumn{3}{c}{\textbf{OPT-175B}} & \multicolumn{3}{c}{\cellcolor[HTML]{D9D9D9}\textbf{text-ada-001}} & \multicolumn{3}{c}{\textbf{text-babbage-001}} & \multicolumn{3}{c}{\cellcolor[HTML]{D9D9D9}\textbf{text-curie-001}} & \multicolumn{3}{c}{\textbf{text-davinci-002}} \\
\multirow{-2}{*}{\textbf{\begin{tabular}[c]{@{}c@{}}Trigger\\ Sentences\end{tabular}}} & \multicolumn{1}{c}{\cellcolor[HTML]{D9D9D9}\textbf{Descr1}} & \multicolumn{1}{c}{\cellcolor[HTML]{D9D9D9}\textbf{Descr2}} & \multicolumn{1}{c}{\cellcolor[HTML]{D9D9D9}\textbf{$\mu$}} & \multicolumn{1}{c}{\textbf{Descr1}} & \multicolumn{1}{c}{\textbf{Descr2}} & \multicolumn{1}{c}{\textbf{$\mu$}} & \multicolumn{1}{c}{\cellcolor[HTML]{D9D9D9}\textbf{Descr1}} & \multicolumn{1}{c}{\cellcolor[HTML]{D9D9D9}\textbf{Descr2}} & \multicolumn{1}{c}{\cellcolor[HTML]{D9D9D9}\textbf{$\mu$}} & \multicolumn{1}{c}{\textbf{Descr1}} & \multicolumn{1}{c}{\textbf{Descr2}} & \multicolumn{1}{c}{\textbf{$\mu$}} & \multicolumn{1}{c}{\cellcolor[HTML]{D9D9D9}\textbf{Descr1}} & \multicolumn{1}{c}{\cellcolor[HTML]{D9D9D9}\textbf{Descr2}} & \multicolumn{1}{c}{\cellcolor[HTML]{D9D9D9}\textbf{$\mu$}} & \multicolumn{1}{c}{\textbf{Descr1}} & \multicolumn{1}{c}{\textbf{Descr2}} & \multicolumn{1}{c}{\textbf{$\mu$}} \\ \midrule
\cellcolor[HTML]{FFFFFF}\textit{1} & 15.57 & 23.35 & 19.46 & 14.97 & 32.93 & 23.95 & 31.14 & 22.16 & 26.65 & 49.10 & 29.94 & 39.52 & 43.11 & 27.54 & 35.33 & \textbf{70.66} & \textbf{59.88} & \textbf{65.27} \\
\textit{2} & 15.57 & 31.14 & 23.35 & 15.57 & 32.93 & 24.25 & 34.13 & 19.16 & 26.65 & 49.10 & 31.14 & 40.12 & 41.32 & 29.94 & 35.63 & 62.87 & 58.08 & 60.48 \\
\cellcolor[HTML]{FFFFFF}\textit{3} & 48.50 & 31.74 & 40.12 & 13.77 & 31.14 & 22.46 & 16.17 & 6.59 & 11.38 & 49.70 & 31.74 & 40.72 & 41.32 & 31.14 & 36.23 & 52.10 & 58.68 & 55.39 \\
\textit{4} & 44.31 & 31.14 & 37.72 & 16.17 & 31.74 & 23.95 & 7.78 & 2.99 & 5.39 & 47.90 & 30.54 & 39.22 & 44.91 & 32.34 & 38.62 & 52.69 & 55.69 & 54.19 \\
\textit{5} & 26.95 & 37.13 & 32.04 & 13.17 & 31.14 & 22.16 & 27.54 & 19.16 & 23.35 & 47.90 & 32.93 & 40.42 & 36.53 & 31.74 & 34.13 & 50.90 & 50.90 & 50.90 \\
\textit{6} & 28.14 & 29.94 & 29.04 & 19.16 & 31.74 & 25.45 & 20.96 & 8.98 & 14.97 & 49.10 & 31.14 & 40.12 & 43.11 & 29.34 & 36.23 & 45.51 & 50.30 & 47.90 \\
\textit{7} & 22.16 & 36.53 & 29.34 & 13.17 & 31.14 & 22.16 & 23.35 & 17.96 & 20.66 & 49.10 & 31.74 & 40.42 & 39.52 & 32.93 & 36.23 & 43.11 & 50.30 & 46.71 \\
\midrule
\multicolumn{1}{l}{\textit{\textbf{Average}}} & 28.74 & 31.57 & 30.15 & 15.14 & 31.82 & 23.48 & 23.01 & 13.86 & 18.43 & 48.85 & 31.31 & 40.08 & 41.40 & 30.71 & 36.06 & 53.98 & 54.83 & 54.41 \\ \bottomrule
\end{tabular}%
}
\end{table*}


\begin{table}[]
\centering
\caption{Breakdown of accuracy of the davinci model using trigger sentence \#1 for buggy event identification and bug type classification (in \%).}
\label{tab:results_davinci}
\resizebox{\columnwidth}{!}{%
\begin{tabular}{@{}lcrrrrrr@{}}
\toprule
 & \multicolumn{1}{l}{} & \multicolumn{3}{c}{\textbf{Buggy Event Identification}} & \multicolumn{3}{c}{\textbf{Bug Type Classification}} \\ 
\multicolumn{1}{c}{\textbf{Game}} & \# & \multicolumn{1}{c}{\textbf{Desc. 1}} & \multicolumn{1}{c}{\textbf{Desc. 2}} & \multicolumn{1}{c}{\textbf{$\mu$}} & \multicolumn{1}{c}{\textbf{Desc. 1}} & \multicolumn{1}{c}{\textbf{Desc. 2}} & \multicolumn{1}{c}{\textbf{$\mu$}} \\ \midrule
\textit{Fallout 4} & 19 & \textbf{89.47} & 68.42 & \textbf{78.95} & 36.84 & 47.37 & 42.11 \\
\textit{Cyberpunk} & 23 & 60.87 & \textbf{82.61} & 71.74 & 47.83 & 39.13 & 43.48 \\
\textit{GTA V} & 21 & 80.95 & 61.90 & 71.43 & 28.57 & 47.62 & 38.10 \\
\textit{Witcher 3} & 21 & 76.19 & 57.14 & 66.67 & 42.86 & 61.90 & 52.38 \\
\textit{Far Cry 5} & 20 & 65.00 & 60.00 & 62.50 & \textbf{55.00} & \textbf{65.00} & \textbf{60.00} \\
\textit{Just Cause 3} & 17 & 58.82 & 64.71 & 61.77 & 41.18 & 52.94 & 47.06 \\
\textit{Skyrim} & 20 & 65.00 & 50.00 & 57.50 & 35.00 & 45.00 & 40.00 \\
\textit{RDR 2} & 26 & 69.23 & 38.46 & 53.85 & 34.62 & 30.77 & 32.70 \\ \bottomrule
\end{tabular}%
}
\end{table}


\paragraph{Robustness of Bug Detection:}
The top performing model for buggy event identification is the davinci model, hence we focus on this model while analyzing the robustness of bug detection.
The Wilcoxon signed rank test shows that only trigger sentence \#1 leads to statistically significant differences between the different sets of descriptions.
Therefore, the model is fairly robust with buggy event identification under most trigger sentences. However, sometimes the chosen set of descriptions can affect the accuracy.

\subsection{Task 2: Bug Type Classification}
We only report the performance of the davinci model for the bug type classification task, as it was the best-performing model in the buggy event identification task. Table~\ref{tab:bugtype_accuracy} shows the bug classification accuracy per bug type. Our results show that the davinci model can correctly predict the bug type 44.01\% of the time for our set of labelled videos. The main source of misclassification is the \textit{Irregular force} bug type, which is also the most frequent bug type in our dataset. 
This type of bug is most often confused with the \textit{Collision}, \textit{Graphics}, \textit{Animation}, and \textit{Teleportation} bug types, which are all game physics bugs (with the exception of \textit{Graphics}). 


The accuracy of the model also varies considerably across games. 
Table~\ref{tab:results_davinci} shows the bug type classification accuracy per game.
The davinci model performs best on the \textit{Fallout 4} game.
The model performs worst on the \textit{Red Dead Redemption 2} game, and this poor performance correlates with the high proportion of \textit{Irregular force} bugs (32 out of 52) for this game in our dataset.

\section{Discussion of Failure Cases}
Here, we discuss cases in which the davinci model failed to identify or reason about the buggy event correctly. 
In each sample box, the ground truth label is green, red shows the model's wrong prediction, and the bold text shows the trigger sentence and answer extraction prompt.

\paragraph{Wrong Reasoning and Wrong Prediction:} 
For some questions, the model fails to complete the intermediate reasoning steps, with outputs being vague or wrong, leading to wrong answers that lack justification. Sample \ref{sample:wrongreasoning} shows such a case of wrong reasoning leading to a wrong prediction for a bug\footnote{\url{https://redd.it/8bspct}} in the \textit{Fallout 4} game.








\begin{figure}[!t]
\begin{davinci}{sample:wrongreasoning}
\small{Q: In the Fallout 4 video game, the following sequence of events happened:

\begin{enumerate}[label=(\alph*)]
\item  A super-mutant shoots a bullet at the player
\item The bullet collides with the player
\hlitemw The player dies
\hlitemc The player walks backwards
\end{enumerate}

\noindent Which event is a bug?

\noindent A:\textbf{Let's reason the events according to the reference game}\llmoutput{ Fallout 4:}

\noindent \llmoutput{(a) A super-mutant shoots a bullet at the player}

\noindent \llmoutput{(b) The bullet collides with the player}

\noindent \llmoutput{(c) The player dies}

\noindent \llmoutput{The player dying is the bug here, as they should not die from a single bullet.}

\noindent\textbf{Among (a) through (d), the answer is}\llmoutput{ (c).}

}
\end{davinci}
\end{figure}


\paragraph{Reasoning Does Not Match the Prediction:} Sometimes, the model's prediction is correct, but the reasoning does not match the prediction. The model follows (some) proper steps toward the (correct) answer, but fails to justify it. 
In Sample \ref{sample:flawed} for the \textit{GTA V} game\footnote{\url{https://redd.it/87lfze}}, the model follows the events and explains the \textit{ragdoll} state correctly. The problem is that the model does not justify its prediction, as it does not explain why the predicted event is the bug.

\begin{figure}[!t]
\begin{davinci}{sample:flawed}
\small{Q: In the Grand Theft Auto V video game, the following sequence of events happened:

\begin{enumerate}[label=(\alph*)]
 \item  A person is driving a car on a freeway
 \item  The person jumps out of the vehicle
 \hlitemc  The person is hit by another car and ejected into the sky
 \item  The person falls from the sky and hits the beach
\end{enumerate}

\noindent Which event is a bug?

\noindent A:\textbf{Let's reason the events according to the reference game}\llmoutput{'s mechanics. In GTA V, when a player exits a vehicle while it's still in motion, the character will automatically enter a ``ragdol'' state and fall to the ground. If the player is then hit by another vehicle while in this ragdoll state, they will be ejected into the air. So event (c) is the bug here.}
}
\end{davinci}
\end{figure}



\paragraph{Wrong Assumptions:} In a few cases, the steps outputted by the model are correct, but one or some assumptions about video games are incorrect. For example, in Sample \ref{sample:correct_reasoning_wrong_answer} for the \textit{Far Cry 5} game\footnote{\url{https://redd.it/88gjyp}}  the davinci model attributes a bug event to a possible earthquake that is not present in the game.

\begin{figure}[!t]
\begin{davinci}{sample:correct_reasoning_wrong_answer}
\small{
Q: In the Far Cry 5 video game, the following sequence of events happened:

\begin{enumerate}[label=(\alph*)]
\item A person approaches the exit of a building
\hlitemc The door is shaking violently at the exit
\hlitemw The person is launched into the sky by the door
\item The person falls back to the ground and is killed
\end{enumerate}

\noindent Which event is a bug?

\noindent A:\textbf{Let's reason the events according to the reference game}\llmoutput{'s logic:}

\noindent \llmoutput{(a) A person approaches the exit of a building - this is fine}

\noindent  \llmoutput{(b) The door is shaking violently at the exit - this is probably due to some kind of earthquake or explosion}

\noindent  \llmoutput{(c) The person is launched into the sky by the door - this is a bug, because there is no in-game logic that would explain why the door would suddenly launch the person into the sky}

\noindent \llmoutput{(d) The person falls back to the ground and is killed - this is fine, because it is a natural consequence of being launched into the sky and then falling back down to the ground}
}

\noindent\textbf{Among (a) through (d), the answer is}\llmoutput{ (c).}
\end{davinci}
\end{figure}

\paragraph{Problem with Reference Game:} We use the \textit{``The reference game is''} trigger sentence to force the model to compare the events within the context of the desired game. Surprisingly, in some cases, this leads to the model complaining about the absence of reference games. Sample~\ref{sample:referencegame} for the \textit{GTA~V} game\footnote{\url{https://redd.it/2x16in}}  shows an example of the model failing to recognize the provided reference game.

\begin{figure}[!t]

\begin{davinci}{sample:referencegame}
\small{Q: In the Grand Theft Auto V video game, the following sequence of events happened:

\begin{enumerate}[label=(\alph*)]
 \item  A bus is on a road
 \hlitemc  The bus is rotating about a vertical axis on the road
 \hlitemw  The rotating bus collides with a person
 \item  The person is knocked to the ground
\end{enumerate}

\noindent Which event is a bug?

\noindent A:\textbf{The reference game is}\llmoutput{ not provided so it is difficult to say which event is a bug.}

\noindent \textbf{Among (a) through (d), the answer is}\llmoutput{ most likely (c).}
}
\end{davinci}
\end{figure}


\section{Future Research Directions}

\paragraph{Improving the Poor Performance of OPT-175B:} The biggest model in the OPT family achieves a low accuracy, even compared to a smaller model of the same family, OPT-66B. Manual review of the output shows that the answer extraction prompt is not suitable for this model. Often the intermediate outputs of the model are correct, but in the second stage, the model outputs a wrong choice as the final answer. Sample~\ref{sample:poorperformance}  demonstrates that the model has correct reasoning for a bug\footnote{\url{https://redd.it/41xesk}} in the \textit{Skyrim} game, but fails in the answer extraction stage. Future studies should investigate how to resolve such issues with answer extraction.


\begin{figure}[!t]
\begin{optxlarge}{sample:poorperformance}
\small{Q: In the The Elder Scrolls V - Skyrim video game, the following sequence of events happened:

\begin{enumerate}[label=(\alph*)]
\item A person stands in front of a monster
\hlitemc The monster rotates and clips through the ground
\item The monster falls off the cliff
\hlitemw The person jumps off the cliff and follows the monster
\end{enumerate}

\noindent Which event is a bug?

\noindent A:\textbf{Let's reason the events according to the reference game}\llmoutput{.}

\noindent \llmoutput{(a) A person stands in front of a monster}

\noindent \llmoutput{This is a normal event.}

\noindent \llmoutput{(b) The monster rotates and clips through the ground}
 
\noindent \llmoutput{This is a bug.}

\noindent \llmoutput{(c) The monster falls off the cliff}

\noindent \llmoutput{This is a normal event.}

\noindent \llmoutput{(d) The person jumps off the cliff and follows the monster}

\noindent \llmoutput{This is a normal event.}

\noindent \llmoutput{The answer is (d).}

\noindent\textbf{Among (a) through (d), the answer is}\llmoutput{ (d).}
}
\end{optxlarge}
\end{figure}






\paragraph{Automating the Description of Event Sequences:}Our primary focus is reasoning about the events in a video game to detect bugs, regardless of the source of the events' descriptions. While it is possible to incorporate a vision model to summarize the events in the video game, this information could also be obtained directly from a game engine. A game engine can be modified to produce both graphical and textual output, which can then be used for bug detection. However, during a preliminary study, we tested CLIP-Cap \cite{mokady2021clipcap}, ZeroCap \cite{tewel2022zerocap} and OFA \cite{wang2022ofa} to create descriptions of videos, and found that none of them can describe frames from video games properly.
Future studies should investigate how the description of event sequences can be automated. 

\paragraph{Searching for an Optimal $Q$:}
In our dataset, it is possible to adapt the question template $Q$ as long as the sequence $S$ is not changed (in contrast to other Q\&A datasets which come with a predefined question template).
Future studies should search for optimal $Q$ formats that lead to higher accuracy. This process requires a held-out set for validating the results to avoid possible biases. As we are interested in assessing the inherent properties of language models in bug detection, we did not fine-tune $Q$ in our experiments, but instead designed a general format for all queries and video games.

\paragraph{Consideration Regarding Inference's Speed:}

As the number of parameters in a model increases, its memory requirement and inference time also grow. Traditional techniques like layer removal~\cite{anwar2017structured, zandigohar2021netcut} and distillation~\cite{hinton2015distilling} could handle this problem, but the performance may suffer. That said, ~\citet{dettmers2022llm} showed using a two-step quantization method, it is possible to reduce the memory footprint of a large model by $2$x without performance degradation.








\section{Conclusion}
In this study, we demonstrated the promising capabilities of language models as video game bug detectors. We introduced a novel dataset of 334 question-answer pairs to extensively evaluate large language models across the OPT and InstructGPT families. Our results are promising but indicate there are challenges in incorporating language models for video game bug detection. In summary, we have provided a new out-of-distribution task for benchmarking large language models, and  we hope to motivate researchers from both the AI and software engineering communities to further explore this promising and exciting new direction of automated  video game testing. 

\bibliography{aaai23}


    
    







\end{document}